\documentclass{article}
\usepackage{ijcai24}

\usepackage{times}
\usepackage{soul}
\usepackage{url}
\usepackage[hidelinks]{hyperref}
\usepackage[utf8]{inputenc}
\usepackage[small]{caption}
\usepackage{graphicx}
\usepackage{amsmath}
\usepackage{amsthm}
\usepackage{booktabs}
\usepackage{algorithm}
\usepackage{algorithmic}
\usepackage[switch]{lineno}
\usepackage{amssymb}
\usepackage{multirow}
\usepackage{graphicx}
\usepackage{subfigure}

\usepackage{subcaption}

\usepackage{dblfloatfix}

\title{InjectTST: A Transformer Method of Injecting Global Information into Independent Channels for Long Time Series Forecasting}

\author{
Ce Chi$^1$\and
Xing Wang$^1$\and
Kexin Yang$^1$\and
Zhiyan Song$^1$\and
Di Jin$^{1}$\and
Lin Zhu$^1$\and\\
Chao Deng$^1$\And
Junlan Feng$^1$
\\
\affiliations
$^1$China Mobile Research Institute\\
\emails
\{chice, wangxing, yangkexin, songzhiyan, jindi, zhulinyj, dengchao, fengjunlan\}@chinamobile.com
}

\begin{document}

\maketitle

\begin{abstract}
    Transformer has become one of the most popular architectures for multivariate time series (MTS) forecasting.
    Recent Transformer-based MTS models generally prefer channel-independent structures with the observation that channel independence can alleviate noise and distribution drift issues, leading to more robustness.
    Nevertheless, it is essential to note that channel dependency remains an inherent characteristic of MTS, carrying valuable information.
    Designing a model that incorporates merits of both channel-independent and channel-mixing structures is a key to further improvement of MTS forecasting, which poses a challenging conundrum.
    To address the problem, an injection method for global information into channel-independent Transformer, InjectTST, is proposed in this paper.
    Instead of designing a channel-mixing model directly, we retain the channel-independent backbone and gradually inject global information into individual channels in a selective way.
    A channel identifier, a global mixing module and a self-contextual attention module are devised in InjectTST.
    The channel identifier can help Transformer distinguish channels for better representation.
    The global mixing module produces cross-channel global information.
    Through the self-contextual attention module, the independent channels can selectively concentrate on useful global information without robustness degradation, and channel mixing is achieved implicitly. 
    Experiments indicate that InjectTST can achieve stable improvement compared with state-of-the-art models.
\end{abstract}

\section{Introduction}

Unprecedented advances have been made in multivariate time series (MTS) forecasting with the rapid growth of deep learning \cite{benidis2022deep}, benefiting various regions, such as finance \cite{patton2013copula}, weather \cite{angryk2020multivariate}, traffic \cite{cai2020traffic}, and networking \cite{liu2021st}.
Recently, Transformer-based MTS models have been widely explored \cite{wen2022transformers} and gradually become one of the most popular architectures for MTS modeling \cite{cao2023tempo,chang2023llm4ts,zhang2023multi}. 

\begin{figure}[htbp]
\centering
\begin{subfigure}{}
    \includegraphics[width=0.45\textwidth]{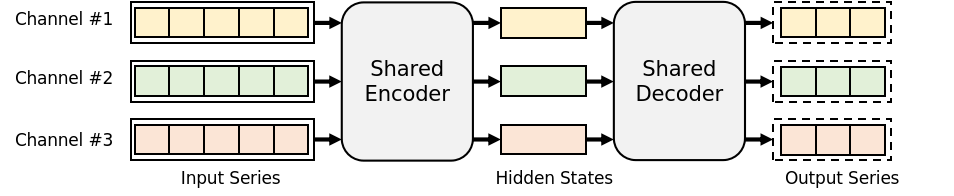}
    \caption*{(a) Channel-independent MTS forecasting model framework.}
\end{subfigure}
\vspace{5mm}
\begin{subfigure}{}
    \includegraphics[width=0.45\textwidth]{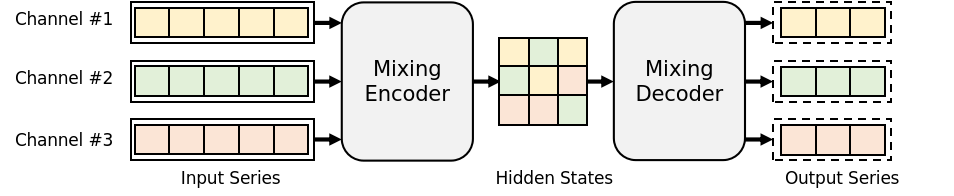}
    \caption*{(b) Channel-mixing MTS forecasting model framework.}
\end{subfigure}
\vspace{1mm}
\begin{subfigure}{}
    \includegraphics[width=0.5\textwidth]{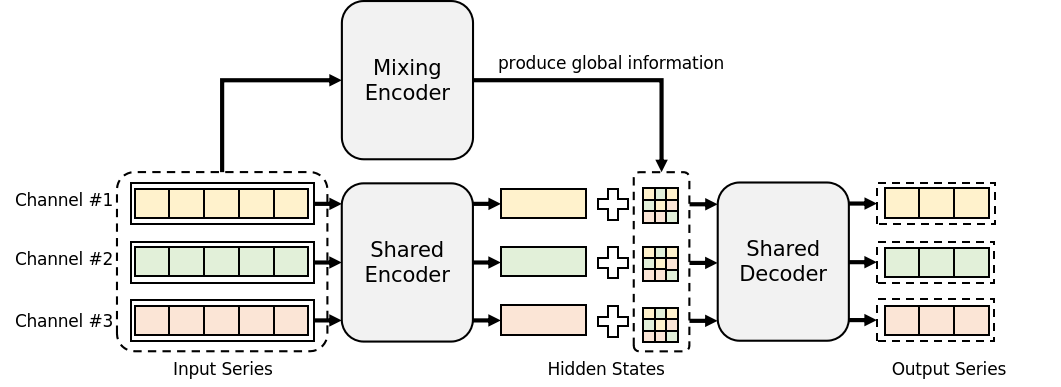}
    \caption*{(c) InjectTST MTS forecasting model framework.}
\end{subfigure}
\caption{Different types of MTS forecasting frameworks. The decoder can be replaced by a simple prediction head. (a) In a channel-independent framework, the prediction of a channel is irrelevant to other channels. The channels share the same model. (b) In a channel-mixing framework, the channels are mixed for a unified representation, and then the decoder produces the prediction for all channels at the same time. (c) In our proposed InjectTST framework, the channel-independent structure is used as a backbone. Each channel receives additional global information so as to achieve channel-mixing implicitly.}
\label{fig:motivation}
\end{figure}

However, recent MTS forecasting models generally prefer a channel-independent structure \cite{nie2022time,cao2023tempo,zhang2023multi,zhou2023one}, as the channel independence usually demonstrates superior performance than channel-mixing structures \cite{nie2022time,han2023capacity}. 
In detail, channel independence (as shown in Fig.~\ref{fig:motivation}(a)) has two merits.
(1) \textbf{noise mitigation}: channel-independent models can focus on individual channel forecasting without being disturbed by noise from other channels \cite{nie2022time}.
(2) \textbf{distribution drift mitigation}: channel independence can alleviate the distribution drift problem of MTS \cite{han2023capacity}.
In contrast, channel mixing proves less effective in dealing with the channel noise and distribution drift issues, leading to inferior performance.
Nevertheless, channel mixing (as shown in Fig~\ref{fig:motivation}(b)) possesses some unique advantages.
(1) \textbf{high information capacity}: channel-mixing models excel in capturing channel dependencies and can bring more information to the forecasting
\cite{ijcai2019p0264,zhang2022crossformer}.
(2) \textbf{channel specificity}: the optimization of multiple channels in channel-mixing models is carried out simultaneously, enabling the model to fully capture the distinct characteristics of each channel. 
In contrast, as channel independence treats channels with a shared model, the model cannot distinguish the channels and mainly learns the general patterns of multiple channels, leading to the loss of channel specificity and potential impact on forecasting.
Hence, designing an effective model with merits of both channel independence and channel mixing, i.e., noise mitigation, distribution drift mitigation, high information capacity and channel specificity, is a key to further enhancing the MTS forecasting performance.

However, designing a model that embodies the merits of both channel independence and channel mixing poses a challenging conundrum.
First, from the perspective of channel independence, channel-independent models are inherently contradictory to channel dependencies.
Although fine-tuning the shared model for each channel can solve the problem of channel specificity, it comes at the expense of substantial training costs.
Second, from the perspective of channel mixing, existing denoising methods and distribution drift-solving methods still struggle to make channel-mixing frameworks as robust as channel independence.
How to transfer the strong noise and distribution drift mitigation capabilities of channel independence to channel-mixing frameworks has not been investigated in previous works.

In order to solve these problems, an MTS injection method for global information into individual channels, InjectTST, is proposed in this paper.
As shown in Fig.~\ref{fig:motivation}(c), in sharp contrast to existing channel-mixing works \cite{wu2022timesnet,zhang2022crossformer}, we avoid to explicitly model the channel dependencies for forecasting.
Instead, our solution is retaining the channel-independent structure as a backbone, and global (channel-mixing) information is injected into each channel selectively so as to achieve channel mixing in an implicit way.
Each individual channel can selectively receive useful global information and avoid noisy one, so high information capacity and noise mitigation can be compatibly maintained.
As channel independence is retained as the backbone, the distribution drift can be alleviated as well.
We introduce a channel identifier into InjectTST to solve the channel specificity problem, which is elaborated in the following.

Specifically, as shown in Fig.~\ref{fig:architecture}, a channel identifier, a global mixing module and a self-contextual attention (SCA) module are devised in InjectTST.
(1) First, we assign each channel with a trainable embedding, named channel identifier.
From the perspective of channels, the channel identifier represents the distinct features of each channel after optimization, which can facilitate the model capability of extracting unique representation of channels.
Moreover, the channel identifier serves the crucial function of distinguishing channels for the Transformer in the injection period, which is similar to positional encoding.
(2) Second, a global mixing module is designed to produce global information for subsequent injection.
Two kinds of global mixing module design are presented in this paper, where a Transformer encoder is used for a high-level global representation.
(3) Third, a self-contextual attention module is designed for harmless information injection.
In our design, the obtained global information is viewed as context, where the injection is achieved via a modified cross attention design.
By selectively concentrating on valuable global information, each channel absorbs the global information with minimal noise disturb.
Hence, channel mixing is implicitly achieved while the robustness is not ruined.
Experiments based on widely used real-world datasets demonstrate that InjectTST can achieves state-of-the-art performance compared with the latest MTS forecasting methods.
More importantly, InjectTST reveals a promising combination solution for MTS modeling, bridging the gap between channel-mixing and channel-independent models.

The contributions of this paper are concluded as follows:

\begin{itemize}
    \item A MTS modeling framework with merits of both channel mixing and channel independence is proposed. In this framework, channel-independent structures are used as backbones, and channel-mixing information is viewed as context and is injected into individual channels in a selective way.
    \item We propose InjectTST, an injection MTS forecasting method for global information into channel-independent Transformer models. In InjectTST, a channel identifier is proposed to identify each channel for better representation. Two kinds of global mixing modules are proposed, which can mix channel information effectively. Viewing global information as context, a cross attention-based SCA module is proposed so as to inject valuable global information into individual channels.
    \item Experimental results show that InjectTST can achieve state-of-the-art in multiple datasets. Extensive analysis indicates that InjectTST is a promising framework for future improvement of both channel-mixing and channel-independent MTS models.
\end{itemize}

\section{Related Works}
Transformer has shown great potential in MTS forecasting. 
In Informer \cite{zhou2021informer}, the long tail distribution issue in MTS Transformer is observed, and a ProbSparse self-attention mechanism is proposed to improve Transformer efficiency.
Autoformer \cite{wu2021autoformer} utilizes MTS decomposition modules to capture the complicated temporal pattern, where an Auto-Correlation mechanism is proposed.
FEDformer \cite{zhou2022fedformer} utilizes decomposition modules as global profile extractors and frequency-based Transformer as detail extractors, where a linear-complexity model is prposed for MTS forecasting.
In Triformer \cite{ijcai2022p0277}, the authors propose a triangular
structure to optimize the Transformer computation complexity, with which a variable-specific parameter scheme is proposed to improve the accuracy of Transformer models.
In Crossformer \cite{zhang2022crossformer}, a two-stage attention mechanism is proposed to capture the cross-channel dependency of MTS.
However, Crossformer is a pure channel-mixing framework, and thus still faces the risk of robustness loss.
In PatchTST \cite{nie2022time}, local semantic information is emphasized and a patching mechanism is proposed.
Besides, channel independence is found effective for MTS forecasting. 
However, channel independence ignores the channel dependency information, so there is still room for improvement.
iTransformer \cite{liu2023itransformer} applies Transformer on inverted dimensions of MTS, where the time series in a channel is viewed as a token.
Therefore, iTransformer focuses on capturing the channel dependency of MTS, and also may result in loss of robustness.

Apart from Transformer-based methods, linear models also show comparable performance for MTS forecasting.
DLinear \cite{zeng2023transformers} applies a simple linear model with decomposition modules and is validated to be effective for MTS forecasting.
TSMixer \cite{ekambaram2023tsmixer} introduces the patching mechanism into MLP-based MTS frameworks, with which an online reconciliation approach is proposed to improve the learning capacity for MLP models.
In FreTS \cite{yi2023frequency}, the authors explore a novel combination of MLP architectures and frequency domain for time series forecasting.
The frequency domain provides a complete view of global dependencies and can make the model concentrate on key parts of frequency components in time series.
However, the contradiction between channel independence and channel mixing remains unresolved in these works.


\section{Methodology}

\begin{figure}[h]
\centerline{\includegraphics[width=0.6\textwidth]{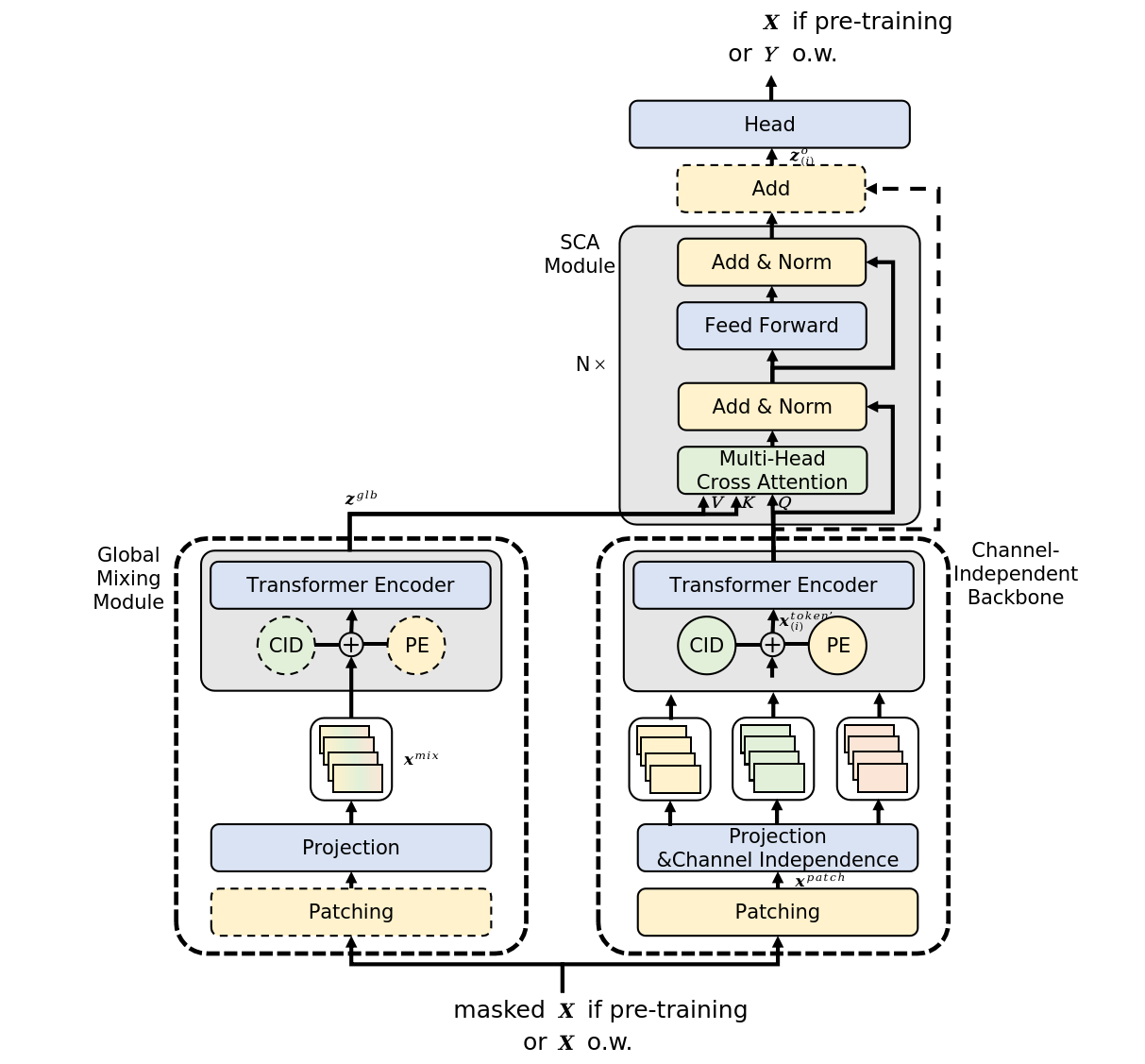}}
\caption{InjectTST architecture. In the channel-independent backbone, the patches of a channel are added with a positional encoding as well as a channel identifier (CID for short). In the global mixing module, the channels are mixed for the global information. Finally, in the self-contextual attention (SCA) module, the global information is injected into each channel via a cross attention design.}
\label{fig:architecture}
\end{figure}

We consider the MTS forecasting problem.
The input is historical $L$ time steps of MTS $\boldsymbol{X} =(\boldsymbol{x}_1,\boldsymbol{x}_2,...,\boldsymbol{x}_L)$, where each time step is a vector with dimension of $M$, i.e.,  $\boldsymbol{x}_t=(x_{t,1},x_{t,2},...,x_{t, M})$, $t=1,2,...,L$.
The target is to forecast the values in future $T$ time steps based on the given history $\boldsymbol{X}$, i.e., $\boldsymbol{Y} = (\boldsymbol{x}_{L+1},...,\boldsymbol{x}_{L+T})$.
To solve the problem and incorporate merits of both channel mixing and independence, InjectTST is proposed in this paper. The architecture of InjectTST is illustrated in Fig.~\ref{fig:architecture}.

\subsection{Channel-Independent Backbone}

\begin{figure*}[b]
\begin{minipage}{0.5\textwidth}
  \centering
  \vspace{8mm}
  \includegraphics[width=0.85\textwidth]{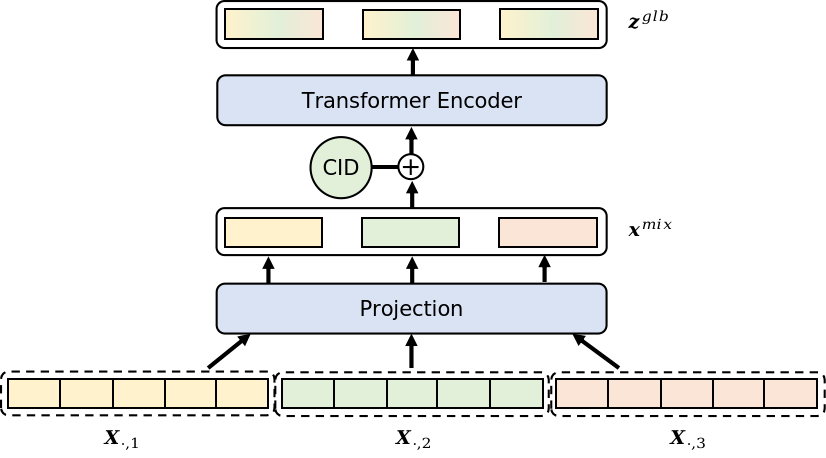}
  \vspace{2mm}
  \caption*{(a) CaT global mixing module.}
  \label{Fig1}
\end{minipage}
\begin{minipage}{0.5\textwidth}
  \centering
  \includegraphics[width=0.5\textwidth]{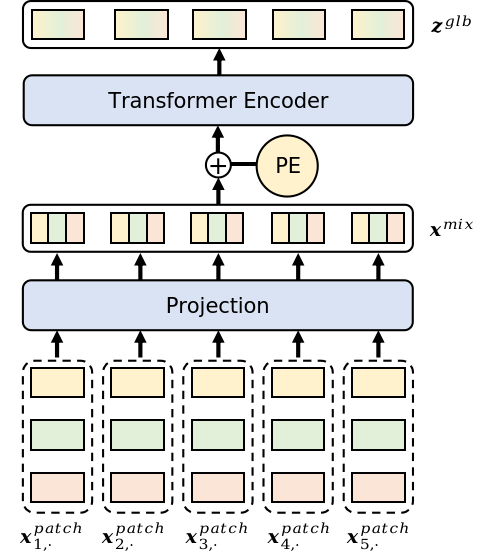}
  \caption*{(b) PaT global mixing module.}
  \label{Fig2}
  \end{minipage}
\caption{Two types of design for the global mixing module.  (a) CaT global mixing module directly projects each channel into tokens, and utilizes a Transformer encoder for cross-channel global mixing. (b) PaT global mixing module groups the patches in the same position and applies a linear projection for local mixing. Then a Transformer encoder is used for global mixing.}
\label{fig:mix}
\end{figure*}

\subsubsection{Patching and Projection}

Patching mechanism is proposed in \cite{nie2022time}, which can improve the  local semantic information and reduce calculation complexity.
We adopt the patching mechanism in InjectTST.
Specifically, in the channel-independent backbone, the input $\boldsymbol{X}\in\mathbb{R}^{L\times M}$ passes through the patching module and produces patches $\boldsymbol{x}^{patch}\in\mathbb{R}^{M\times PN \times PL}$, where $PL$ denotes the length of each patch and $PN$ denotes the number of patches in each channel. $PN=\lfloor \frac{L-PL}{S} \rfloor+2$, where $S$ denotes the stride of patching.
After that, a linear projection $\boldsymbol{W}\in\mathbb{R}^{PL\times D}$ with a learnable positional encoding $\boldsymbol{U}\in\mathbb{R}^{PN\times D}$ is used to project patches into initial tokens, where $D$ denotes the dimension of the Transformer model.
The process can be represented as 
\begin{equation}
    \boldsymbol{x}^{token} = \boldsymbol{x}^{patch}\boldsymbol{W} + \boldsymbol{U},
\end{equation}
where the output $\boldsymbol{x}^{token} \in \mathbb{R}^{M\times PN\times D}$.

\subsubsection{Channel Identifier}
The channel-independent framework treats channels with a shared model. 
As a result, the model cannot distinguish the channels and mainly learns general patterns of the channels, lacking channel specificity.


To solve the problem, we introduce channel identifier into InjectTST, which is a learnable tensor $\boldsymbol{V}\in\mathbb{R}^{M\times D}$.
After the linear projection of patches, the tokens are added with both positional encoding and channel identifier, i.e., 
\begin{equation}
\boldsymbol{x}^{token\prime}=\boldsymbol{x}^{token} + \boldsymbol{V} = \boldsymbol{x}^{patch}\boldsymbol{W} + \boldsymbol{U}+\boldsymbol{V}.
\end{equation}

Finally, $\boldsymbol{x}^{token\prime}$ is input into the Transformer encoder for high-level representation in a channel-independent way:
\begin{equation}
    \boldsymbol{z}_{(i)} = \text{Encoder}^{\text{ci}}(\boldsymbol{x}^{token\prime}_{(i)}), i=1,2,...,M.
\end{equation}

The channel identifier represents the distinct features of each channel, which enables the model to distinguish channels and capture unique representation of channels.

\subsection{Global Mixing Module}


In the channel-mixing route, the input sequence $\boldsymbol{X}$ first pass through a global mixing module to produce global information.
The major target of InjectTST is to inject global information into each channel, so how to obtain global information is a critical problem.
Two kinds of effective global mixing modules are presented in this paper, named CaT (channel as a token) and PaT (patch as a token) respectively.

\subsubsection{CaT Global Mixing Module}
Specifically, as shown in Fig.~\ref{fig:mix}(a), the CaT module directly projects each channel into a token.
In detail, a linear project $\boldsymbol{W}_{mix}\in\mathbb{R}^{L \times D}$ is applied on the whole channel values:
\begin{equation}
    \boldsymbol{x}^{mix} = \boldsymbol{X}^{\text{T}}\boldsymbol{W}_{mix},
\end{equation}
where $\boldsymbol{x}^{mix}\in\mathbb{R}^{M\times D}$.
Then $\boldsymbol{x}^{mix}$ is input into a Transformer encoder with our proposed channel identifier $\boldsymbol{V}$ to produce the final global information:
\begin{equation}
    \boldsymbol{z}^{glb} = \text{Encoder}^{\text{mix}}(\boldsymbol{x}^{mix} + \boldsymbol{V}).
\end{equation}

\subsubsection{PaT Global Mixing Module}
PaT global mixing module is shown in Fig.~\ref{fig:mix}(b).
PaT takes patches $\boldsymbol{x}^{patch}$ as inputs.
It first reshapes the patches into dimensions of $PN\times(M\cdot PL)$ so that the patches in the same relative position are grouped.
Then, a linear projection $\boldsymbol{W}_{mix}\in\mathbb{R}^{(M\cdot PL) \times D}$ is applied on the grouped patches, i.e., 
\begin{equation}
    \boldsymbol{x}^{mix} = \boldsymbol{x}^{patch}\boldsymbol{W}_{mix},
\end{equation}
where $\boldsymbol{x}^{mix}\in\mathbb{R}^{PN\times D}$.
The linear projection mainly fuses information within the patch level. 
After that, a Transformer encoder is applied on $\boldsymbol{x}^{mix}$ with positional encoding $\boldsymbol{U}$ to further fuse information cross patches, and the global information $\boldsymbol{z}^{glb}$ is obtained:
\begin{equation}
    \boldsymbol{z}^{glb} = \text{Encoder}^{\text{mix}}(\boldsymbol{x}^{mix} + \boldsymbol{U}).
\end{equation}

Experiments illustrate that PaT is more stable while CaT is outstanding in some special datasets.
Details are presented in Section~\ref{sec:exp_global_mixing}.



\subsection{Self-Contextual Attention Module}
One of the challenges in our framework is that we must inject global information into each channel with minimal impact on the robustness.
In the vanilla Transformer, cross attention makes the target sequence selectively and freely focus on context information from another source according relevance.
With this insight, the cross attention design could be suitable for the MTS injection as well.

To be specific, the global information mixed from channels $\boldsymbol{z}^{glb}$ can be viewed as context.
We use a cross attention design to inject the global information into each channel, as shown in Fig.\ref{fig:architecture}.
A residual connection is optional in our design, which is marked with dotted lines in the SCA module in Fig.\ref{fig:architecture}.
In general cases, the residual connection makes the model slightly unstable.
However, as depicted in Section~\ref{sec:exp_rc}, the residual connection can achieve outstanding improvement in some special datasets.

Overall, the global information $\boldsymbol{z}^{glb}$ is input into the SCA module as keys and values.
And the channel information $\boldsymbol{z}_{(i)}$ is input as queries:
\begin{equation}
    \boldsymbol{z}^{o}_{(i)} = \text{SCA}(\boldsymbol{z}_{(i)}, \boldsymbol{z}^{glb}, \boldsymbol{z}^{glb}).
\end{equation}

After the cross attention, the output $\boldsymbol{z}^{o}_{(i)}$ is improved by global information.
Finally, a linear head is appended to produce the prediction.
We adopt a self-supervised training method same as \cite{nie2022time}.
In the pre-training stage, the head is a pre-training head, which predicts the masked patches.
In the finetuning stage, the head is a prediction head, which predicts the future series.


\subsection{Self-Supervised Training and Normalization}

\begin{table*}[b]
\centering
\begin{tabular}{c|c|cc|cc|cc|cc|cc|cc|cc}
\hline
\multicolumn{2}{c|}{Models} & \multicolumn{2}{c|}{InjectTST} & \multicolumn{2}{c|}{PatchTST} & \multicolumn{2}{c|}{iTransformer} & \multicolumn{2}{c|}{DLinear} & \multicolumn{2}{c|}{TimesNet} & \multicolumn{2}{c|}{FEDformer} & \multicolumn{2}{c}{Autoformer} \\
\hline
\multicolumn{2}{c|}{Metric} & MSE & MAE & MSE & MAE & MSE & MAE & MSE & MAE & MSE & MAE & MSE & MAE & MSE & MAE \\
\hline
\multirow{4}{*}{\rotatebox[origin=c]{90}{Weather}} 
&  96  & \textbf{0.140} & \textbf{0.192} & 0.144 & 0.193 & 0.176 & 0.216 & 0.176 & 0.237 & 0.170 & 0.220 & 0.238 & 0.314 & 0.249 & 0.329 \\
& 192  & \textbf{0.184} & \textbf{0.235} & 0.193 & 0.239 & 0.225 & 0.257 & 0.220 & 0.282 & 0.224 & 0.263 & 0.275 & 0.329 & 0.325 & 0.370 \\
& 336  & \textbf{0.236} & \textbf{0.278} & 0.244 & 0.280 & 0.281 & 0.299 & 0.265 & 0.319 & 0.282 & 0.304 & 0.339 & 0.377 & 0.351 & 0.391 \\
& 720  & \textbf{0.308} & \textbf{0.331} & 0.318 & 0.334 & 0.358 & 0.350 & 0.323 & 0.362 & 0.357 & 0.353 & 0.389 & 0.409 & 0.415 & 0.426 \\
\hline
\multirow{4}{*}{\rotatebox[origin=c]{90}{Traffic}} 
&  96  & \textbf{0.360} & \textbf{0.250} & 0.376 & 0.261 & 0.393 & 0.268 & 0.410 & 0.282 & 0.589 & 0.315 & 0.576 & 0.359 & 0.597 & 0.371 \\
& 192  & \textbf{0.387} & \textbf{0.259} & 0.390 & 0.267 & 0.413 & 0.277 & 0.423 & 0.287 & 0.618 & 0.324 & 0.610 & 0.607 & 0.382 & 0.2 \\
& 336  & \textbf{0.399} & \textbf{0.265} & \textbf{0.399} & 0.271 & 0.425 & 0.283 & 0.436 & 0.296 & 0.635 & 0.341 & 0.608 & 0.375 & 0.623 & 0.387 \\
& 720  & 0.442 & 0.303 & \textbf{0.436} & \textbf{0.293} & 0.458 & 0.300 & 0.466 & 0.315 & 0.664 & 0.352 & 0.621 & 0.375 & 0.639 & 0.395 \\
\hline
\multirow{4}{*}{\rotatebox[origin=c]{90}{Electricity}} 
&  96  & \textbf{0.125} & \textbf{0.219} & 0.126 & 0.221 & 0.148 & 0.239 & 0.140 & 0.237 & 0.168 & 0.271 & 0.186 & 0.302 & 0.196 & 0.313 \\
& 192  & \textbf{0.145} & 0.239 & \textbf{0.145} & \textbf{0.238} & 0.167 & 0.258 & 0.153 & 0.249 & 0.191 & 0.293 & 0.197 & 0.311 & 0.211 & 0.324 \\
& 336  & \textbf{0.161} & \textbf{0.256} & 0.164 & \textbf{0.256} & 0.179 & 0.272 & 0.169 & 0.267 & 0.197 & 0.299 & 0.213 & 0.328 & 0.214 & 0.327 \\
& 720  & \textbf{0.198} & 0.291 &\textbf{ 0.198} & \textbf{0.289} & 0.208 & 0.297 & 0.203 & 0.301 & 0.262 & 0.345 & 0.233 & 0.344 & 0.236 & 0.342 \\
\hline
\multirow{4}{*}{\rotatebox[origin=c]{90}{ETTh1}} 
&  96  & 0.367 & \textbf{0.394} & \textbf{0.366} & 0.397 & 0.393 & 0.408 & 0.375 & 0.399 & 0.389 & 0.412 & 0.376 & 0.415 & 0.435 & 0.446 \\
& 192  & \textbf{0.400} & \textbf{0.415} & 0.401 & 0.416 & 0.446 & 0.440 & 0.405 & 0.416 & 0.440 & 0.442 & 0.423 & 0.446 & 0.456 & 0.457 \\
& 336  & \textbf{0.432} & \textbf{0.438} & 0.450 & 0.456 & 0.494 & 0.467 & 0.439 & 0.443 & 0.495 & 0.471 & 0.444 & 0.462 & 0.486 & 0.487 \\
& 720  & \textbf{0.468} & \textbf{0.482} & 0.472 & 0.484 & 0.517 & 0.501 & 0.472 & 0.490 & 0.518 & 0.495 & 0.469 & 0.492 & 0.515 & 0.517 \\
\hline
\multirow{4}{*}{\rotatebox[origin=c]{90}{ETTh2}} 
&  96  & \textbf{0.275} & \textbf{0.337} & 0.284 & 0.343 & 0.301 & 0.350 & 0.289 & 0.353 & 0.332 & 0.370 & 0.332 & 0.374 & 0.332 & 0.368 \\
& 192  & \textbf{0.344} & \textbf{0.385} & 0.357 & 0.387 & 0.380 & 0.399 & 0.383 & 0.418 & 0.397 & 0.410 & 0.407 & 0.446 & 0.426 & 0.434 \\
& 336  & \textbf{0.370} & \textbf{0.406} & 0.377 & 0.410 & 0.424 & 0.432 & 0.448 & 0.465 & 0.453 & 0.451 & 0.400 & 0.447 & 0.477 & 0.479 \\
& 720  & \textbf{0.398} & 0.436 & 0.400 & \textbf{0.435} & 0.430 & 0.447 & 0.605 & 0.551 & 0.438 & 0.450 & 0.412 & 0.469 & 0.453 & 0.490 \\
\hline
\multirow{4}{*}{\rotatebox[origin=c]{90}{ETTm1}} 
&  96  & \textbf{0.285} & \textbf{0.343} & 0.289 & 0.344 & 0.342 & 0.377 & 0.299 & 0.343 & 0.335 & 0.377 & 0.326 & 0.390 & 0.510 & 0.492 \\
& 192  & \textbf{0.323} & \textbf{0.365} & 0.330 & 0.369 & 0.383 & 0.396 & 0.335 & \textbf{0.365} & 0.405 & 0.411 & 0.365 & 0.415 & 0.514 & 0.495 \\
& 336  & 0.354 & \textbf{0.383} & \textbf{0.353} & 0.387 & 0.418 & 0.418 & 0.369 & 0.386 & 0.416 & 0.422 & 0.392 & 0.425 & 0.510 & 0.492 \\
& 720  & \textbf{0.403} & \textbf{0.413} & 0.407 & 0.415 & 0.487 & 0.457 & 0.425 & 0.421 & 0.479 & 0.459 & 0.446 & 0.458 & 0.527 & 0.493 \\
\hline
\multirow{4}{*}{\rotatebox[origin=c]{90}{ETTm2}} 
&  96  & \textbf{0.165} & \textbf{0.254} & 0.170 & 0.257 & 0.186 & 0.272 & 0.167 & 0.260 & 0.188 & 0.266 & 0.180 & 0.271 & 0.205 & 0.293 \\
& 192  & \textbf{0.219} & \textbf{0.293} & 0.221 & 0.295 & 0.254 & 0.314 & 0.224 & 0.303 & 0.263 & 0.311 & 0.252 & 0.318 & 0.278 & 0.336 \\
& 336  & \textbf{0.268} & \textbf{0.324} & 0.269 & 0.325 & 0.316 & 0.351 & 0.281 & 0.342 & 0.322 & 0.349 & 0.324 & 0.364 & 0.343 & 0.379 \\
& 720  & \textbf{0.354} & \textbf{0.382} & 0.365 & 0.388 & 0.414 & 0.407 & 0.397 & 0.421 & 0.424 & 0.408 & 0.410 & 0.420 & 0.414 & 0.419 \\
\hline
\end{tabular}
\caption{Multivariate time series forecasting results of various models. InjectTST refers to InjectTST-PaT for specification. PatchTST here is the self-supervised version which generally outperforms its supervised one so as to present a strong baseline. Prediction length is set in \{96, 192, 336, 720\}. The best results are in bold.}
\label{tab:comparison}
\end{table*}

The training process consists of 3 stages.
First, in the pre-training stage, the input time series are randomly masked and the target is to predict the masked parts.
Second, in the head-finetuning stage, the pre-training head of InjectTST is replaced by a prediction head and we finetune the prediction head while the rest of the network is frozen.
Third, in the finetuning stage, the entire network of InjectTST is finetuned.

Moreover, we adopt a simple normalization introduced in DLinear\cite{zeng2023transformers,han2023capacity} to further alleviate the distribution drift problem.
The method modifies the input by subtracting the last values of the input sequence. After prediction, the last values are added back to the prediction.
Therefore, the prediction is the difference from the latest value, and the distribution drift problem can be alleviated.

\section{Experiments}
\label{sec:exp}

\subsection{Experimental Setup}
\begin{figure*}[t]
\centerline{\includegraphics[width=0.85\textwidth]{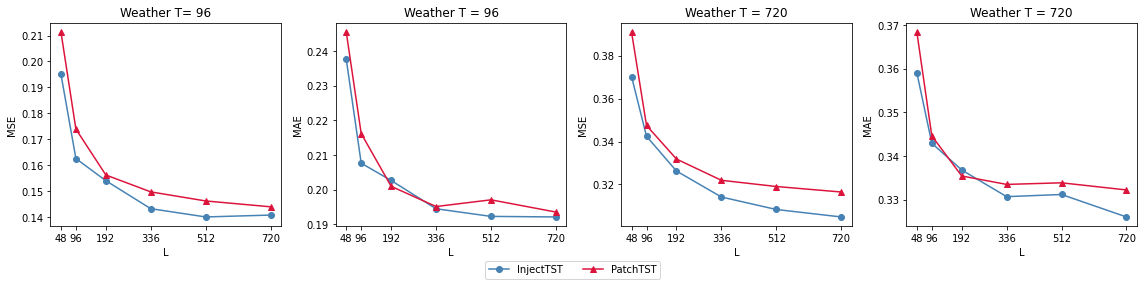}}
\caption{Forecasting performance of InjectTST and PatchTST with varing historical sequence lengths. The historical sequence lengths $L$ are set in \{48, 96, 192, 336, 512, 720\}. The prediction lengths $T$ are selected in \{96, 720\}.}
\label{fig:varing_history}
\end{figure*}
\subsubsection{Datasets}

Multiple widely used real-world MTS datasets are used in our experiments, including (1) Weather, (2) Electricity, (3) Traffic, (4) ETTh1, (5) ETTh2, (6) ETTm1, (7) ETTm2. Please refer to \cite{wu2021autoformer} for dataset details.

\subsubsection{Baselines}
Various well known latest MTS forecasting baselines are compared, including (1) Transformer-based models: PatchTST \cite{nie2022time}, iTransformer,
\cite{liu2023itransformer}, Autoformer \cite{wu2021autoformer}, FEDformer \cite{zhou2022fedformer}; 
(2) a MLP-based model: DLinear \cite{zeng2023transformers};
(3) a CNN-based model: TimesNet \cite{wu2022timesnet}.

\subsubsection{Implementation Details}
Our Experiments are conducted in a server equipped with Intel(R) Xeon(R) Gold 5118 CPU @ 2.30GHz, 376 GB of RAM, and 2 Tesla V100-PCIE-32GB GPUs.
The experiment is implemented with PyTorch 1.11.0 \cite{paszke2019pytorch} based on the code framework of PatchTST \cite{nie2022time}.

The input historical sequence length is set to 512.
The patch length is set to 12 and stride is set to 12.
The mask ratio of self-supervised learning is set to 50\%.
There are 3 steps for the training. First, we pre-train InjectTST for 20 epochs. Second, we finetune the prediction head of InjectTST for 10 epochs. Finally, the entire network of InjectTST is finetuned for 100 epochs.
Mean Square Error (MSE) and Mean Absolute Error (MAE) are used as evaluation metrics.

\subsubsection{Model Variants}
As different kinds of global mixing modules are designed and the residual connection in the SCA module is optional, InjectTST is a set of models, including InjectTST-CaT, InjectTST-CaT-RC, InjectTST-PaT, InjectTST-PaT-RC, respectively.
RC means the model adopts the optional residual connection in SCA.
In the main results, InjectTST-PaT is used as the major model, and unless otherwise specified, InjectTST refers to InjectTST-PaT.

\subsection{Main Results}

As shown in Table~\ref{tab:comparison}, InjectTST achieves SOTA in most of datasets.
The improvement in Weather, ETTh1, ETTh2 and ETTm2 is significant.
On average, InjectTST achieves 1.63\% reduction on MSE and 0.89\% on MAE compared with PatchTST, one of the strongest baselines.
It indicates that InjectTST can effectively inject global information into individual channels and improve the forecasting performance with very little impact on the robustness of channel-independent structures.

\subsection{Ablation Study}
\subsubsection{Effect of Each Component}

The importance of the channel identifier and global information is analyzed in Table.~\ref{tab:wo-compotents}.
The global mixing and SCA modules are closely related, so we directly present a model without global information, named InjectTST w/o GI.
It can be seen that the model performance degrades significantly without global information, indicating the absolute importance of global information in InjectTST.

On the other hand, according to the results of PatchtTST and InjectTST w/o GI, where the channel identifier plays a major role, it can be observed that the channel identifier is also effective for MTS forecasting.
Although InjectTST w/o CID achieves similar performance to InjectTST, the stability becomes inferior without the channel identifier.
In summary, the channel identifier plays a role in making the overall performance more stable and has a slight positive effect on performance improvement.

\subsubsection{Varying Historical Sequence Lengths}

The performance of InjectTST with different historical sequence lengths is tested.
As shown in Fig.~\ref{fig:varing_history}, with the benefits of the patching mechanism, the performance of InjectTST can be improved with the increase of input length.
Moreover, InjectTST can generally outperform PatchTST with different input lengths, validating the effectiveness of InjectTST.

\subsection{Analysis}


\begin{table*}[htpb]
\centering
\begin{tabular}{c|c|cc|cc|cc|cc}
\hline
\multicolumn{2}{c|}{Models} & \multicolumn{2}{c|}{InjectTST} & \multicolumn{2}{c|}{InjectTST w/o CID} & \multicolumn{2}{c|}{InjectTST w/o GI} & \multicolumn{2}{c}{PatchTST} \\
\hline
\multicolumn{2}{c|}{Metric} & MSE & MAE & MSE & MAE & MSE & MAE & MSE & MAE \\
\hline
\multirow{5}{*}{\rotatebox[origin=c]{90}{Weather}}  
& 96    & \textbf{0.140 } & \textbf{0.192 } & 0.144  & 0.194  & \underline{0.143}  & 0.197  & 0.144  & \underline{0.193}  \\
& 192   & \textbf{0.184 } & \textbf{0.235 } & \underline{0.186}  & \underline{0.238}  & \underline{0.186}  & \underline{0.238}  & 0.193  & 0.239  \\
& 336   & \textbf{0.236 } & \underline{0.278}  & \textbf{0.236 } & \textbf{0.275 } & 0.237  & 0.281  & 0.244  & 0.280  \\
& 720   & \textbf{0.308 } & \underline{0.331}  & \textbf{0.308 } & \textbf{0.328 } & 0.316  & 0.335  & 0.318  & 0.334  \\
& Avg   & \textbf{0.217}  & \textbf{0.259}  & \underline{0.218}  & \textbf{0.259}  & 0.220  & 0.263  & 0.225  & 0.262 \\
\hline
\multirow{5}{*}{\rotatebox[origin=c]{90}{ETTh2}}  
& 96    & \underline{0.275}  & \underline{0.337}  & \textbf{0.274 } & \textbf{0.335 } & 0.276  & 0.338  & 0.284  & 0.343  \\
& 192   & \underline{0.344}  & \textbf{0.385 } & \textbf{0.343 } & \underline{0.386}  & 0.351  & \underline{0.386}  & 0.357  & 0.387  \\
& 336   & 0.370  & 0.406  & \underline{0.369}  & \textbf{0.403 } & \textbf{0.365 } & \underline{0.405}  & 0.377  & 0.410  \\
& 720   & \textbf{0.398 } & 0.436  & 0.402  & \textbf{0.434 } & 0.420  & 0.455  & \underline{0.400}  & \underline{0.435}  \\
& Avg   & \textbf{0.347}  & \underline{0.391}  & \textbf{0.347}  & \textbf{0.389}  & 0.353  & 0.396  & 0.354  & 0.394  \\
\hline
\multirow{5}{*}{\rotatebox[origin=c]{90}{ETTm1}}  
& 96    & 0.285  & 0.343  & \textbf{0.283 } & \textbf{0.338 } & \underline{0.284}  & \underline{0.340}  & 0.289  & 0.344  \\
& 192   & \textbf{0.323 } & \underline{0.365}  & 0.326  & \textbf{0.363 } & \underline{0.325}  & 0.367  & 0.330  & 0.369  \\
& 336   & \underline{0.354}  & \textbf{0.383 } & \underline{0.354}  & \underline{0.385}  & 0.357  & 0.387  & \textbf{0.353} & 0.387 \\
& 720   & \textbf{0.403 } & \textbf{0.413 } & \underline{0.406}  & 0.419  & \underline{0.406}  & 0.417  & 0.407  & \underline{0.415}  \\
& Avg   & \textbf{0.341}  & \textbf{0.376}  & \underline{0.342}  & \textbf{0.376}  & 0.343  & 0.378  & 0.345  & 0.379  \\
\hline
\end{tabular}
\caption{Ablation Study of channel identifier (CID) and global information (GI). InjectTST w/o CID means that the channel identifier is removed from InjectTST, so global information plays a major role. InjectTST w/o GI means that the global information, including the global mixing module and self-contextual attention module, is removed from InjectTST, so the channel identifier plays a major role. The best results are marked in bold and the second best are underlined.}
\label{tab:wo-compotents}
\end{table*}
\begin{table*}[htpb]
\centering
\begin{tabular}{c|c|cc|cc|cc|cc|cc}
\hline
\multicolumn{2}{c|}{Models} 
& \multicolumn{2}{c|}{InjectTST} 
& \multicolumn{2}{c|}{InjectTST-CaT} 
& \multicolumn{2}{c|}{InjectTST-PaT-RC} 
& \multicolumn{2}{c|}{InjectTST-CaT-RC} 
& \multicolumn{2}{c}{PatchTST} \\
\hline
\multicolumn{2}{c|}{Metric} & MSE & MAE & MSE & MAE & MSE & MAE & MSE & MAE & MSE & MAE \\
\hline
\multirow{5}{*}{\rotatebox[origin=c]{90}{Weather}} 
& 96    & \textbf{0.140}  & \textbf{0.192}  & 0.142  & 0.194  & 0.141  & \underline{0.193}  & \underline{0.141}  & 0.194  & 0.144  & \underline{0.193}  \\
& 192   & \textbf{0.184}  & \underline{0.235}  & 0.186  & \textbf{0.234}  & 0.186  & 0.239  & \textbf{0.184}  & 0.236  & 0.193  & 0.239  \\
& 336   & 0.236  & \underline{0.278}  & \textbf{0.235}  & \textbf{0.275}  & 0.240  & 0.279  & \textbf{0.235}  & 0.278  & 0.244  & 0.280  \\
& 720   & \textbf{0.308}  & \underline{0.331}  & \underline{0.310}  & \textbf{0.329}  & 0.320  & 0.339  & 0.311  & 0.333  & 0.318  & 0.334  \\
& Avg   & \textbf{0.217}  & \underline{0.259}  & \underline{0.218}  & \textbf{0.258}  & 0.222  & 0.263  & \underline{0.218}  & 0.260  & 0.225  & 0.262  \\
\hline
\multirow{5}{*}{\rotatebox[origin=c]{90}{ETTh1}}
& 96    & 0.367  & 0.394  & \textbf{0.360}  & \textbf{0.392}  & \underline{0.365}  & 0.394  & \underline{0.365}  & \underline{0.393}  & 0.366  & 0.397  \\
& 192   & \underline{0.400}  & \textbf{0.415}  & \textbf{0.396}  & \underline{0.416}  & 0.401  & 0.417  & 0.406  & 0.418  & 0.401  & \underline{0.416}  \\
& 336   & 0.432  & 0.438  & 0.455  & 0.462  & \textbf{0.422}  & \underline{0.429}  & \underline{0.423}  & \textbf{0.428}  & 0.450  & 0.456  \\
& 720   & 0.468  & 0.482  & 0.465  & 0.479  & \underline{0.439}  & \underline{0.453}  & \textbf{0.426}  & \textbf{0.444}  & 0.472  & 0.484  \\
& Avg   & 0.417  & 0.432  & 0.419  & 0.437  & \underline{0.407}  & \underline{0.423}  & \textbf{0.405}  & \textbf{0.421}  & 0.422  & 0.438  \\
\hline
\multirow{5}{*}{\rotatebox[origin=c]{90}{ETTh2}}
& 96    & 0.275  & \underline{0.337}  & 0.273  & \textbf{0.336}  & \underline{0.272}  & 0.338  & \textbf{0.271}  & 0.337  & 0.284  & 0.343  \\
& 192   & 0.344  & 0.385  & 0.343  & 0.387  & \textbf{0.336}  & \textbf{0.381}  & \textbf{0.336}  & \underline{0.382}  & 0.357  & 0.387  \\
& 336   & 0.370  & 0.406  & \underline{0.367}  & \textbf{0.403}  & \textbf{0.364}  & 0.407  & \underline{0.367}  & \underline{0.405}  & 0.377  & 0.410  \\
& 720   & \underline{0.398}  & 0.436  & \textbf{0.396}  & \textbf{0.434}  & 0.414  & 0.448  & 0.409  & 0.445  & 0.400  & \underline{0.435} \\
& Avg   & 0.347  & \underline{0.391}  & \textbf{0.345}  & \textbf{0.390}  & 0.347  & 0.394  & \underline{0.346}  & 0.392  & 0.354  & 0.394  \\
\hline
\multirow{5}{*}{\rotatebox[origin=c]{90}{ETTm1}}
& 96    & 0.285  & 0.343  & \underline{0.282}  & \underline{0.338}  & \underline{0.282}  & 0.341  & \textbf{0.280}  & \textbf{0.337}  & 0.289  & 0.344  \\
& 192   & \textbf{0.323}  & \textbf{0.365}  & \textbf{0.323}  & \textbf{0.365}  & 0.326  & 0.367  & \textbf{0.323}  & 0.366  & 0.330  & 0.369  \\
& 336   & \underline{0.354}  & \textbf{0.383}  & 0.355  & 0.386  & 0.356  & 0.386  & \underline{0.354}  & \underline{0.385}  & \textbf{0.353} & 0.387 \\
& 720   & \textbf{0.403}  & \textbf{0.413}  & 0.407  & 0.418  & 0.414  & 0.422  & \underline{0.404}  & 0.416  & 0.407  & \underline{0.415}  \\
& Avg   & \underline{0.341}  & \textbf{0.376}  & 0.342  & 0.377  & 0.345  & 0.379  & \textbf{0.340}  & \textbf{0.376}  & 0.345  & 0.379  \\
\hline
\multirow{5}{*}{\rotatebox[origin=c]{90}{ETTm2}}
& 96    & 0.165  & \textbf{0.254}  & 0.166  & 0.256  & \textbf{0.164}  & \textbf{0.254}  & \textbf{0.164}  & \textbf{0.254}  & 0.170  & 0.257  \\
& 192   & \textbf{0.219}  & \textbf{0.293}  & 0.222  & 0.296  & 0.220  & \textbf{0.293}  & \textbf{0.219}  & \textbf{0.293}  & 0.221  & 0.295  \\
& 336   & \textbf{0.268}  & \textbf{0.324}  & 0.272  & 0.329  & 0.276  & 0.332  & \underline{0.269}  & \underline{0.325}  & \underline{0.269}  & \underline{0.325}  \\
& 720   & \underline{0.354}  & \textbf{0.382}  & 0.357  & 0.386  & \textbf{0.351}  & \textbf{0.382}  & 0.356  & 0.388  & 0.365  & 0.388  \\
& Avg   & \textbf{0.251}  & \textbf{0.313}  & 0.254  & 0.317  & 0.253  & \underline{0.315}  & \underline{0.252}  & \underline{0.315}  & 0.256  & 0.316  \\
\hline
\end{tabular}
\caption{Variants of InjectTST. InjectTST refers to InjectTST-PaT for specification, which utilizes PaT global mixing design. InjectTST-CaT adopts a CaT global mixing design. InjectTST-PaT-RC and InjectTST-CaT-RC add an additional residual connection in the self-contextual attention module. The best results are in bold and the second best are underlined.}
\label{tab:rc}
\end{table*}

\subsubsection{Different Design of Global Mixing Module}
\label{sec:exp_global_mixing}
We change the component design of InjectTST to further validate the effectiveness of the InjectTST framework.
First, we change the global mixing module from PaT to CaT, as shown in Table.~\ref{tab:rc}.
It can be seen that InjectTST-CaT can achieve comparable improvement compared with InjectTST.
Although InjectTST-CaT shows worse stability, the improvement of InjectTST-CaT compared with PatchTST can still validate the effectiveness of the injection framework design.



\subsubsection{Effect of Residual Connection}
\label{sec:exp_rc}

An optional residual connection in SCA can be added so that the selection freedom of channels can be further improved.
We try this design and the results are presented in Table.~\ref{tab:rc}.
Two observations are worth noting.
First, InjectTST-PaT-RC and InjectTST-CaT-RC achieve much better performance compared with ones without RC in the ETTh1 dataset.
Second, InjectTST-PaT-RC and InjectTST-CaT-RC become more unstable in the cases of longer series forecasting, such as prediction 720 steps in Weather, ETTh2, and ETTm1 datasets.
The reason may come from the different characteristics of the datasets.
If a dataset needs less channel dependencies, a little global information can make great improvement so that residual connection is needed to mitigate redundant global information.
More importantly, all of the InjectTST variants present considerable improvement.
It indicates that the injection design is effective and potential for MTS forecasting.

\section{Conclusion and Future Work}
Designing a model that can incorporate the merits of both channel independence and channel mixing is a key to further improvement of MTS forecasting.
In this paper, we provide a different way to achieve channel-mixing time series forecasting. That is, channel-independent structures are used as backbones and cross-channel global information is injected into individual channels so as to improve the information content of each channel.
In our proposed InjectTST, a channel identifier, a global mixing module, and a self-contextual attention module are devised to achieve a selective and harmless injection framework.
Experiments demonstrate that InjectTST can effectively improve forecasting performance with negligible impact on robustness. 

InjectTST not only presents an effective MTS forecasting method but also a new framework for channel mixing.
For injection framework design, the global mixing module can be further explored for more effectiveness and efficiency.
The self-contextual attention module for injection can also be replaced by a more effective design so that noise can be futher reduced.
In summary, InjectTST can be a potential framework to bridge channel-independent and channel-mixing models for MTS forecasting.



\clearpage

\bibliographystyle{named}
\bibliography{ijcai24}

\end{document}